\documentclass{article} 
\usepackage{times}
\usepackage[
 letterpaper,
 textheight=9in,
 textwidth=5.5in,
 top=1in
 ]{geometry}
\usepackage{epsfig,graphics}
\usepackage[outercaption]{sidecap}
\usepackage{xcolor}
\usepackage{subfigure}
\usepackage{epstopdf}
\definecolor{mydarkblue}{rgb}{0,0.08,0.45}
\usepackage[bookmarks, colorlinks=true, citecolor=mydarkblue,linkcolor=mydarkblue,urlcolor=mydarkblue]{hyperref}
\usepackage{url}
\usepackage{amsmath,amssymb,amsthm}
\usepackage{mhchem}
\usepackage{authblk}

\newcommand{\presec}{\vspace*{-2pt}}
\newcommand{\postsec}{\vspace*{-2pt}} 

\newcommand{\precap}{\vspace*{-0in}}
\newcommand{\postcap}{\vspace*{-0in}}

\usepackage[ruled]{algorithm2e}

\usepackage[scaled=0.95]{inconsolata}

\title{Training Deep Nets with Sublinear Memory Cost}
\date{}
\author[]{\textbf{Tianqi Chen} $ ^1$}
\author[]{\textbf{Bing Xu} $ ^2$}
\author[]{\textbf{Chiyuan Zhang} $ ^3$}
\author[]{\textbf{Carlos Guestrin} $ ^1$}
\affil[]{$ ^1$ Unveristy of Washington \ \ $ ^2$ Dato. Inc \ \ $ ^3$ Massachusetts Institute of Technology }

\begin{document}
\maketitle
\begin{abstract}
We propose a systematic approach to reduce the memory consumption of deep neural network training.
Specifically, we design an algorithm that costs $O(\sqrt{n})$ memory
to train a $n$ layer network, with only the computational cost of an extra forward pass per
mini-batch.
As many of the state-of-the-art models hit the upper bound of the GPU memory,
our algorithm allows deeper and more complex models to be explored, and helps advance the
innovations in deep learning research.
We focus on reducing the memory cost to store the intermediate feature maps and gradients during training.
Computation graph analysis is used for automatic in-place operation and memory sharing optimizations.
We show that it is possible to trade computation for memory giving a more memory efficient training algorithm with a little extra computation cost.
In the extreme case, our analysis also shows that the memory consumption can be reduced to $O(\log
n)$ with as little as $O(n \log n)$ extra cost for forward computation.
Our experiments show that we can reduce the memory cost of a 1,000-layer deep residual network from
48G to 7G on ImageNet problems. Similarly, significant memory cost
reduction is observed in training complex recurrent neural networks on very long sequences.
\end{abstract}

\presec
\section{Introduction}
\postsec

In this paper, we propose a systematic approach to reduce the memory consumption of deep neural network training.
We mainly focus on reducing the memory cost to store intermediate results~(feature maps) and
gradients, as the size of the parameters are relatively small comparing to the size of
the intermediate feature maps in many common deep architectures.
We use a computation graph analysis to do automatic in-place operation and memory sharing optimizations.
More importantly, we propose a novel method to trade computation for memory.
As a result, we give a practical algorithm that cost $O(\sqrt{n})$ memory for feature maps to train
a $n$ layer network with only double the forward pass computational cost.
Interestingly, we also show that in the extreme case, it is possible to use as little as $O(\log n)$ memory for the features maps to train a $n$ layer network.

We have recently witnessed the success of deep neural networks in many domains~\cite
{Goodfellow-et-al-2016-Book}, such as computer vision, speech recognition, natural language
processing and reinforcement learning.
Many of the success are brought by innovations in new architectures of deep neural networks.
Convolutional neural networks~\cite{lecun-cnn,Alexnet,Ioffe+Szegedy-2015, He2015} model
the spatial patterns and give the state of art results in computer vision tasks.
Recurrent neural networks, such as long short-term memory~\cite{Hochreiter:LSTM},
show inspiring results in sequence modeling and structure prediction.
One common trend in those new models is to use deeper architectures~\cite
{srivastava2015highway,Alexnet,Ioffe+Szegedy-2015, He2015}
to capture the complex patterns in a large amount of training data.
Since the cost of storing feature maps and their gradients scales linearly with the depth of network, our capability of exploring deeper models is limited by the device (usually a GPU) memory.
For example, we already run out of memories in one of the current state-of-art models as described
in \cite{He2016}.
In the long run, an ideal machine learning system should be able to continuously learn from
an increasing amount of training data.
Since the optimal model size and complexity often grows with more training data, it is very
important to have memory-efficient training algorithms.

Reducing memory consumption not only allows us to train bigger models.
It also enables larger batch size for better device utilization and stablity of batchwise operators
such as batch normalization~\cite{Ioffe+Szegedy-2015}.
For memory limited devices, it helps improve memory locality and potentially leads to better memory
access patterns. It also enables us to switch from model parallelism to data parallelism for training
deep convolutional neural networks, which can be beneficial in certain circumstances.
Our solution enables us to train deeper convolutional neural networks, as well as
recurrent neural networks with longer unrolling steps. We provide guidelines for deep learning
frameworks to incorporate the memory optimization techniques proposed in this paper. We will also
make our implementation of memory optimization algorithm publicly available.

\presec
\section{Related Works}\label{sec:rel}
\postsec

We can trace the idea of computational graph and liveness analysis
back to the literatures of compiler optimizations~\cite{Aho:DragonBook}.
Analogy between optimizing a computer program and optimizing a deep neural network computational
graph can be found. For example, memory allocation in deep networks is similar to register
allocation in a compiler.
The formal analysis of computational graph allows us save memory in a principled way.
Theano~\cite{bergstra+al:2010-scipy,Bastien-Theano-2012} is a
pioneering framework to bring the computation graph to deep learning,
which is joined by recently introduced frameworks such as CNTK~\cite{CNTK}, Tensorflow~\cite
{tensorflow2015-whitepaper} and MXNet~\cite{MXNet-whitepaper}.
Theano and Tensorflow use reference count based recycling and runtime garbage collection to manage
memory during training,
while MXNet uses a static memory allocation strategy prior to the actual computation.
However, most of the existing framework focus on graph analysis to optimize computation after the
gradient graph is constructed, but do not discuss the computation and memory trade-off.

The trade-off between memory and computation has been a long standing topic in systems research.
Although not widely known, the idea of dropping intermediate results is also known as gradient checkpointing
technique in automatic differentiation literature~\cite{Griewank:resolve}.
We bring this idea to neural network gradient graph construction for general deep neural networks.
Through the discussion with our colleagues~\cite{Yu:Highway}, we know that the idea of dropping computation has been
applied in some limited specific use-cases.
In this paper, we propose a general methodology that works for general deep neural networks,
including both convolutional and recurrent neural networks.
Our results show that it is possible to train a general deep neural network with sublinear memory
cost.
More importantly, we propose an automatic planning algorithm to provide a good memory
plan for real use-cases.
The proposed gradient graph optimization algorithm can be readily \emph{combined with all the existing
memory optimizations} in the computational graph to further reduce the memory consumption of deep
learning
frameworks.

\begin{figure}
\centering
\includegraphics[width=.98\textwidth]{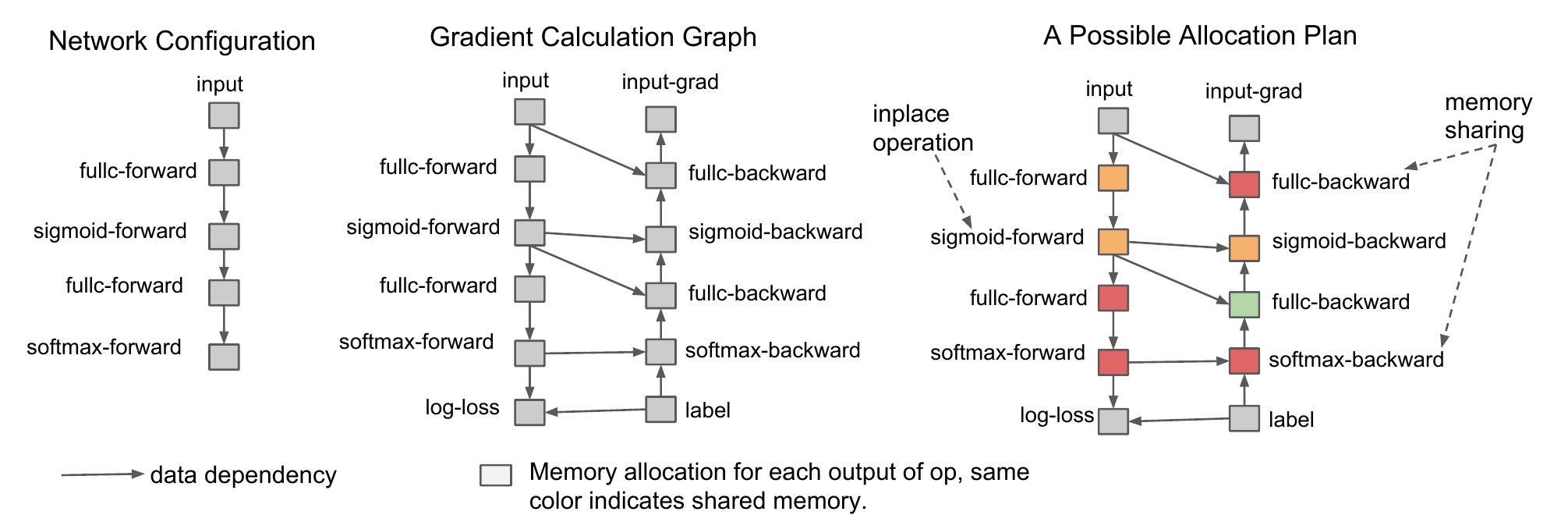}
\precap
\caption{Computation graph and possible memory allocation plan of a two layer fully connected neural network training procedure.
Each node represents an operation and each edge represents a dependency between the operations.
The nodes with the same color share the memory to store output or back-propagated gradient in each operator.
To make the graph more clearly, we omit the weights and their output gradient nodes from the graph and assume that the gradient of weights are also calculated during backward operations. We also annotate two places where the in-place and sharing strategies are used.
}\label{fig:comp-graph}
\postcap
\end{figure}

There are other ways to train big models, such as swapping of CPU/GPU memory and use of model parallel training~\cite{GoogleDisbelief, rhu2016virtualizing}.
These are orthogonal approaches and can be used together with our algorithm to train even bigger models with fewer resources.
Moreover, our algorithm does not need additional communication over PCI-E and can save the bandwidth for model/data parallel training.

\presec
\section{Memory Optimization with Computation Graph}\label{sec:comp-graph}
\postsec

We start by reviewing the concept of computation graph and the memory optimization techniques.
Some of these techniques are already used by existing frameworks such as Theano~\cite{bergstra+al:2010-scipy,Bastien-Theano-2012}, Tensorflow~\cite{tensorflow2015-whitepaper} and MXNet~\cite{MXNet-whitepaper}.
A computation graph consists of operational nodes and edges that represent the dependencies between the operations.
Fig.~\ref{fig:comp-graph} gives an example of the computation graph of a two-layer fully connected
neural network.
Here we use coarse grained forward and backward operations to make the graph simpler.
We further simplify the graph by hiding the weight nodes and gradients of the weights.
A computation graph used in practice can be more complicated and contains mixture of fine/coarse grained operations.
The analysis presented in this paper can be directly used in those more general cases.

Once the network configuration~(forward graph) is given, we can construct the corresponding backward pathway for gradient calculation.
A backward pathway can be constructed by traversing the configuration in reverse topological order, and apply the backward operators as in normal back-propagation algorithm.
The backward pathway in Fig.~\ref{fig:comp-graph} represents the gradient calculation steps \emph{explicitly}, so that the gradient calculation step in training is simplified to just a forward pass on the entire computation graph~(including the gradient calculation pathway).
Explicit gradient path also offers some other benefits~(e.g. being able to calculate higher order gradients), which is beyond our scope and will not be covered in this paper.

When training a deep convolutional/recurrent network, a great proportion of the memory is usually used to store the intermediate outputs and gradients. Each of these intermediate results corresponds
to a node in the graph. A smart allocation algorithm is able to assign the least amount of memory to these nodes
by sharing memory when possible. Fig.~\ref{fig:comp-graph} shows a possible allocation plan of the example two-layer neural network. Two types of memory optimizations can be used
\begin{itemize}
   \item \emph{Inplace operation}: Directly store the output values to memory of a input value.
   \item \emph{Memory sharing}: Memory used by intermediate results that are no longer needed can be recycled and used in another node.
\end{itemize}
Allocation plan in Fig.~\ref{fig:comp-graph} contains examples of both cases. The first sigmoid
transformation is carried out using inplace operation to save memory, which is then reused by its backward
operation.
The storage of the softmax gradient is shared with the gradient by the first fully connected layer.
Ad hoc application of these optimizations can leads to errors.
For example, if the input of an operation is still needed by another operation, applying inplace operation on the input will lead to a wrong result.

\begin{figure}
\centering
\includegraphics[width=.98\textwidth]{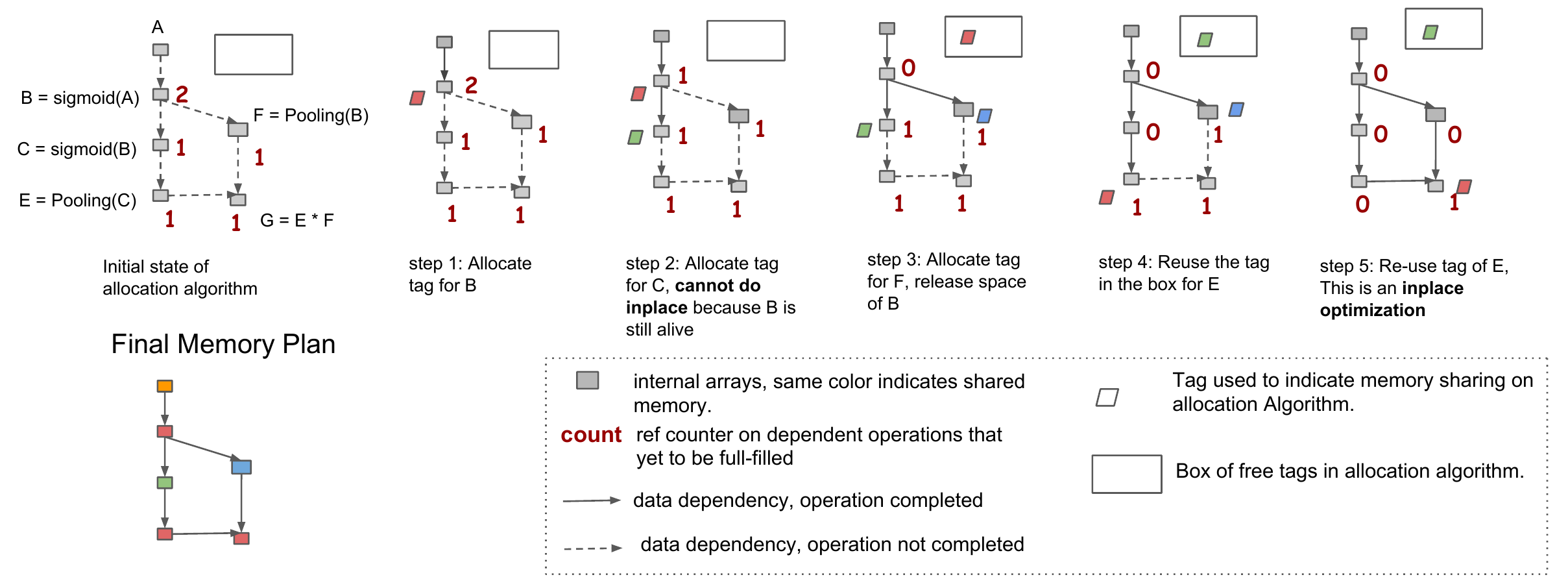}
\precap
\caption{Memory allocation algorithm on computation graph.
Each node associated with a liveness counter to count on operations to be full-filled.
A temporal tag is used to indicate memory sharing.
Inplace operation can be carried out when the current operations is the only one left~(input of counter equals 1).
The tag of a node can be recycled when the node's counter goes to zero.
}\label{fig:mem-alloc}
\postcap
\end{figure}

We can only share memory between the nodes whose lifetime do not overlap.
There are multiple ways to solve this problem.
One option is to construct the conflicting graph of with each variable as node and edges between
variables with overlapping lifespan and then run a graph-coloring algorithm.
This will cost $O(n^2)$ computation time. We adopt a simpler heuristic with only $O(n)$ time. The algorithm is demonstrated in Fig.~\ref{fig:mem-alloc}.
It traverses the graph in topological order, and uses a counter to indicate the liveness of each record.
An inplace operation can happen when there is no other pending operations that depend on its input. Memory sharing happens when a recycled tag is used by another node. This can also serve as a dynamic runtime algorithm that
traverses the graph, and use a garbage collector to recycle the outdated memory. We use this as a static memory allocation algorithm, to allocate the memory to each node before the execution starts, in order to avoid the overhead of garbage collection during runtime.

\noindent \textbf{Guidelines for Deep Learning Frameworks} As we can see from the algorithm demonstration graph in Fig.~\ref{fig:mem-alloc}.
The data dependency causes longer lifespan of each output and increases the memory consumption of big network.
It is important for deep learning frameworks to
\begin{itemize}
   \item Declare the dependency requirements of gradient operators in minimum manner.
   \item Apply liveness analysis on the dependency information and enable memory sharing.
\end{itemize}

It is important to declare minimum dependencies.
For example, the allocation plan in Fig.~\ref{fig:comp-graph} won't be possible if
\texttt{sigmoid-backward} also depend on the output of the first \texttt{fullc-forward}.
The dependency analysis can usually reduce the memory footprint of deep network prediction of a $n$ layer network from $O(n)$ to nearly $O(1)$ because sharing can be done between each intermediate results.
The technique also helps to reduce the memory footprint of training, although only up to a constant factor.

\presec
\section{Trade Computation for Memory}
\postsec

\subsection{General Methodology}

The techniques introduced in Sec.~\ref{sec:comp-graph} can reduce the memory footprint for both training and prediction of deep neural networks.
However, due to the fact that most gradient operators will depend on the intermediate results
of the forward pass, we still need $O(n)$ memory for intermediate results to train a
$n$ layer convolutional network or a recurrent neural networks with a sequence of length $n$.
In order to further reduce the memory, we propose to \emph{drop some of the intermediate results}, and recover them from an extra forward computation when needed.

More specifically, during the backpropagation phase, we can re-compute the dropped intermediate results by running forward from the closest recorded results.
To present the idea more clearly, we show a simplified algorithm for a linear chain feed-forward neural network in Alg.~\ref{alg:backprop}.
Specifically, the neural network is divided into several segments. The algorithm only remembers the
output of each segment and drops all the intermediate results within each segment. The dropped
results are recomputed at the segment level during back-propagation.
As a result, we only need to pay the memory cost to store the outputs of each segment plus the maximum memory cost to do backpropagation on each segment.

\begin{algorithm}[!t]
   \caption{Backpropagation with Data Dropping in a Linear Chain Network}\label{alg:backprop}
     $v \leftarrow input$\\
     \For{$k=1$ {\bfseries to} $length(segments)$}{
        $temp[k] \leftarrow v$\\
        \For{$i=segments[k].begin$ {\bfseries to} $segments[k].end - 1$}{
           $v\leftarrow layer[i].forward(v)$
        }
     }
     $g \leftarrow gradient(v, label)$\\
     \For{$k = length(segments) $ {\bfseries to} $1$}{
        $v \leftarrow temp[k] $\\
        $localtemp \leftarrow \mbox{empty hashtable}$\\
        \For{$i=segments[k].begin$ {\bfseries to} $segments[k].end - 1$}{
           $localtemp[i]\leftarrow v$\\
            $v\leftarrow layer[i].forward(v)$\\
        }
        \For{$i=segments[k].end -1 $ {\bfseries to} $segments[k].begin $}{
           $g \leftarrow layer[i].backward(g, localtemp[i])$
        }
     }
\end{algorithm}

\begin{figure}
\centering
\includegraphics[width=.8\textwidth]{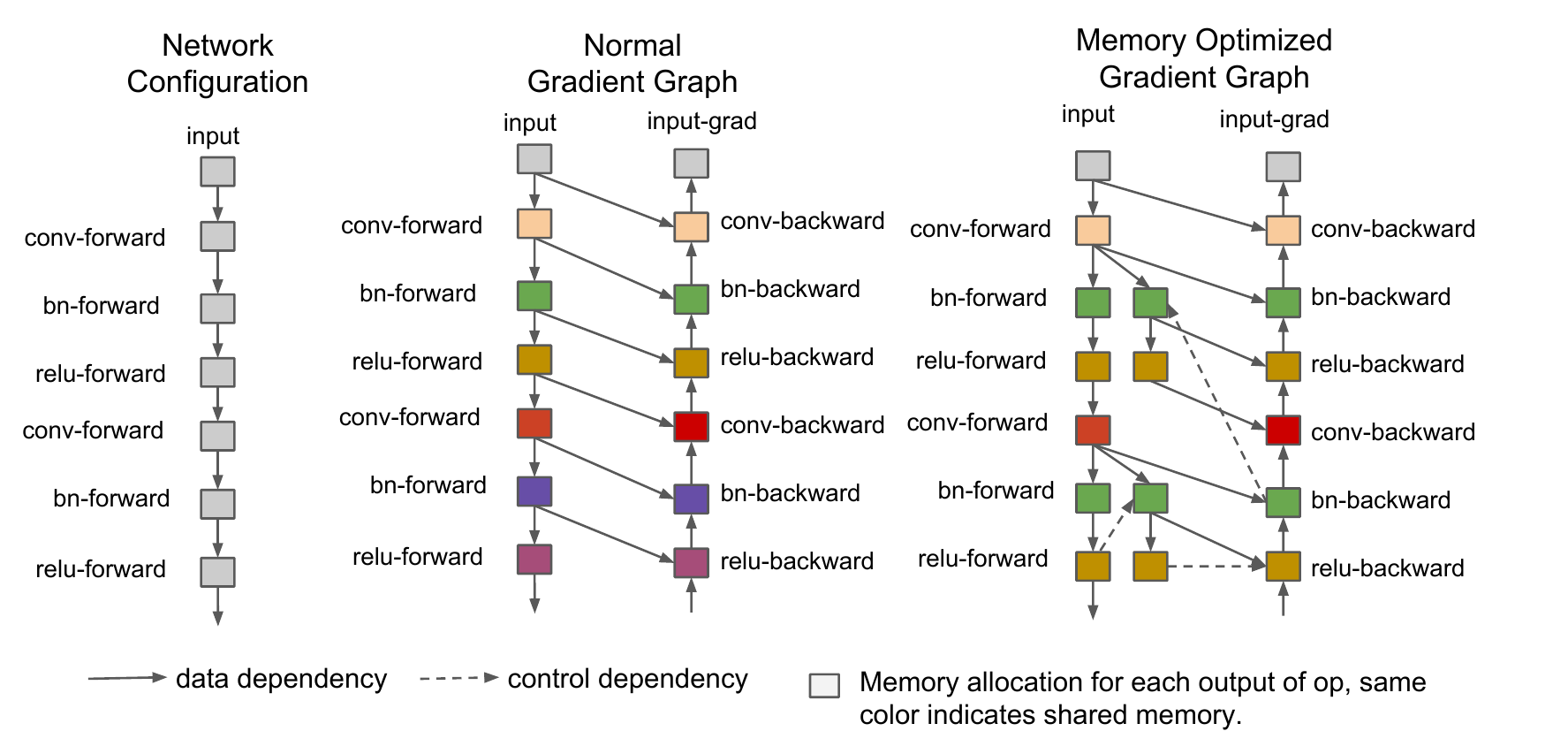}
\precap
\caption{Memory optimized gradient graph generation example.
The forward path is \emph{mirrored} to represent the re-computation happened at gradient calculation.
User specifies the mirror factor to control whether a result should be dropped or kept.
}\label{fig:mirror}
\postcap
\end{figure}

Alg.~\ref{alg:backprop} can also be generalized to common computation graphs as long as we can divide the graph into segments.
However, there are two drawbacks on directly applying Alg.~\ref{alg:backprop}: 1) users have to manually divide the graph and write customized training loop;
2) we cannot benefit from other memory optimizations presented in Sec~\ref{sec:comp-graph}.
We solve this problem by introducing a general gradient graph construction algorithm that uses essentially the same idea.
The algorithm is given in Alg.~\ref{alg:mirror}. In this algorithm, the user specify a function
$m:\mathcal{V}\rightarrow\mathbb{N}$ on the nodes of a computation graph to indicate how many times
a result can be recomputed. We call $m$ the mirror count function as the re-computation is essentially duplicating (mirroring) the nodes.
When all the mirror counts are set to $0$, the algorithm degenerates to normal gradient graph.
To specify re-computation pattern in Alg.~\ref{alg:mirror}, the user only needs to set the $m(v)=1$
for nodes within each segment and $m(v)=0$ for the output node of each segment.
The mirror count can also be larger than 1, which leads to a recursive generalization to be
discussed in Sec~\ref{sec:recursion}.
Fig.~\ref{fig:mirror} shows an example of memory optimized gradient graph. Importantly, Alg.~\ref
{alg:mirror} also outputs a traversal order for the computation, so the memory usage can be
optimized. Moreover, this traversal order can help introduce control flow dependencies for
frameworks that depend on runtime allocation.

\begin{algorithm}[!t]
   \caption{Memory Optimized Gradient Graph Construction}\label{alg:mirror}
   \KwIn{$G=(V, \mbox{pred})$, input computation graph, the $\mbox{pred}[v]$ gives the predecessors array of node $v$.}
   \KwIn{$\mbox{gradient(succ\_grads, output, inputs)}$, symbolic gradient function that creates a gradient node given successor gradients and output and inputs}
   \KwIn{$m: V\rightarrow \mathbb{N}^+$, $m(v)$ gives how many time node $v$ should be duplicated, $m(v)=0$ means do no drop output of node $v$.}
   $ a[v] \leftarrow v \mbox{ for } v \in V$\\
   \For{$k=1$ {\bfseries to} $\max_{v\in V} m(v)$}{
       \For{$v \mbox{ in } \mbox{topological-order}(V)$}{
           \If{$k \leq m(v)$} {
               $a[v] \leftarrow \mbox{new node, same operator as v}$\\
               $\mbox{pred}[a[v]] \leftarrow \cup_{u \in \mbox{pred}[v]}\{ a[u]\}$\\
           }
       }
   }
   $V' \leftarrow \mbox{topological-order}(V)$\\
   \For{$v \mbox{ in } \mbox{reverse-topological-order}(V)$}{
       $g[v] \leftarrow gradient([ g[v] \mbox{ for v in } successor(v)],  a[v], [a[v] \mbox{ for v in } pred[v] ])$\\
       $V' \leftarrow append(V', \mbox{topological-order}(acenstors(g[v])) - V')$
   }
   \KwOut{$G' = (V', \mbox{pred})$ the new graph, the order in $V'$ gives the logical execution order.}
\end{algorithm}

\begin{algorithm}[!t]
   \caption{Memory Planning with Budget}\label{alg:pack}
   \KwIn{$G=(V, \mbox{pred})$, input computation graph.}
   \KwIn{$C \subset V$, candidate stage splitting points, we will search splitting points over $v\subset C$ }
   \KwIn{$B$, approximate memory budget. We can search over $B$ to optimize the memory allocation.}
   $temp \leftarrow 0, x \leftarrow 0, y\leftarrow 0$\\
   \For{$v \mbox{ in } \mbox{topological-order}(V)$}{
       $temp \leftarrow temp + \mbox{size-of-output}(v)$\\
       \uIf{$v\in C$ \mbox{ and  }$temp > B$}{
           $x \leftarrow x + \mbox{size-of-output}(v)$,
           $y \leftarrow max(y, temp)$\\
           $m(v) = 0$, $temp \leftarrow 0$
       }
       \Else{
           $m(v) = 1$\\
       }
   }
   \KwOut{$x$ approximate cost to store inter-stage feature maps}
   \KwOut{$y$ approximate memory cost for each sub stage}
   \KwOut{$m$ the mirror plan to feed to Alg.~\ref{alg:mirror}}
\end{algorithm}

\subsection{Drop the Results of Low Cost Operations}
One quick application of the general methodology is to drop the results of low cost operations and keep the results that are time consuming to compute.
This is usually useful in a \texttt{Conv-BatchNorm-Activation} pipeline in convolutional neural
networks. We can always keep the result of convolution, but drop the result of the batch
normalization, activation function and pooling. In practice this will translate to a memory saving
with little computation overhead, as the computation for both batch normalization and activation
functions are cheap.

\subsection{An $O(\sqrt{n})$ Memory Cost Algorithm}
Alg.~\ref{alg:mirror} provides a general way to trade computation for memory.
It remains to ask which intermediate result we should keep and which ones to re-compute.
Assume we divide the $n$ network into $k$ segments the memory cost to train this network is given as follows.
\begin{equation}
   \mbox{cost-total} =  \max_{i=1,\ldots,k} \mbox{cost-of-segment}(i) + O(k) = O\left(\frac{n}
   {k}\right) + O(k)
\end{equation}
The first part of the equation is the memory cost to run back-propagation on each of the segment.
Given that the segment is equally divided, this translates into $O(n/k)$ cost.
The second part of equation is the cost to store the intermediate outputs between segments.
Setting $k =\sqrt{n}$, we get the cost of $O( 2 \sqrt{n} )$.
This algorithm \emph{only requires an additional forward pass} during training,
but reduces the memory cost to be \emph{sub-linear}.
Since the backward operation is nearly twice as time consuming as the forward one, it only slows down the
computation by a small amount.

In the most general case, the memory cost of each layer is not the same, so we cannot simply
set $k =\sqrt{n}$. However, the trade-off between the intermediate outputs and the cost of each
stage still holds.  In this case, we use Alg.~\ref{alg:pack} to do a greedy allocation with a given
budget for the memory cost within each segment as a single parameter $B$.
Varying $B$ gives us various allocation plans that either assign more memory to the intermediate
outputs, or to computation within each stage.
When we do static memory allocation, we can get the \emph{exact memory cost} given each allocation plan.
We can use this information to do a heuristic search over $B$ to find optimal memory plan that balances the cost of the two.
The details of the searching step is presented in the supplementary material.
We find this approach works well in practice.
We can also generalize this algorithm by considering the cost to run each operation to try to keep time consuming operations when possible.

\subsection{More General View: Recursion and Subroutine}
\label{sec:recursion}

\begin{figure}
\centering
\includegraphics[width=.6\textwidth]{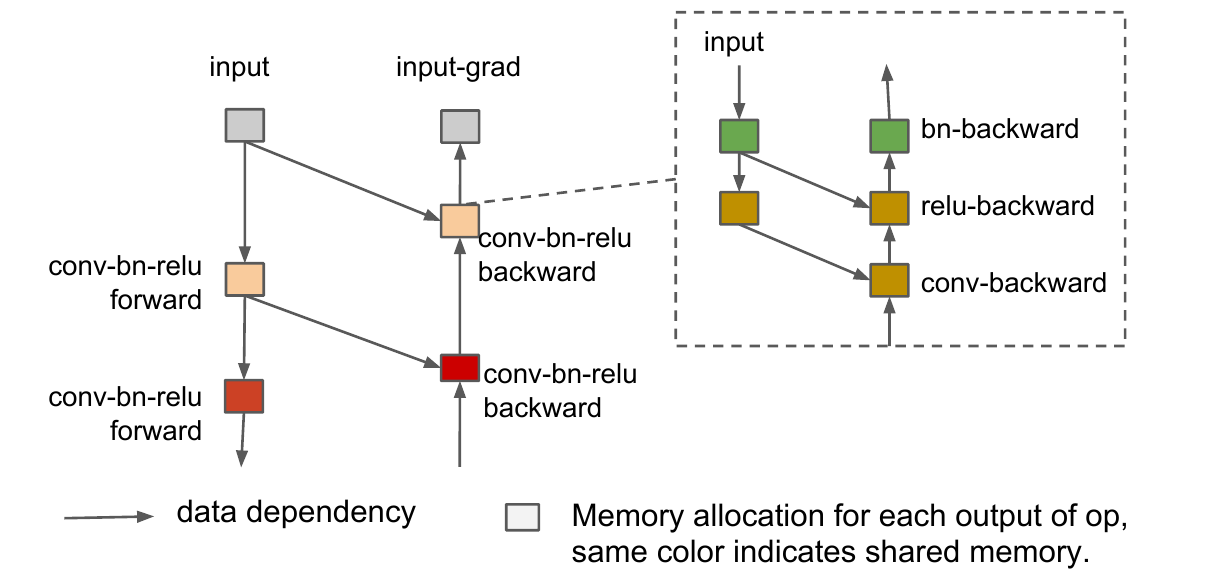}
\precap
\caption{Recursion view of the memory optimized allocations.
The segment can be viewed as a single operator that combines all the operators within the segment.
Inside each operator, a sub-graph as executed to calculate the gradient.
}\label{fig:recursion}
\postcap
\end{figure}

In this section, we provide an alternative view of the memory optimization scheme described above.
Specifically, we can view each segment as a bulk operator that combines all the operations inside the segment together. The idea is illustrated in Fig.~\ref{fig:recursion}.
The combined operator calculates the gradient by executing over the sub-graph that describes
its internal computation.
This view allows us to treat a series of operations as subroutines. The optimization within the
sub-graph does not affect the external world.
As a result, we can recursively apply our memory optimization scheme to each sub-graph.

\paragraph{Pay Even Less Memory with Recursion}
Let $g(n)$ to be the memory cost to do forward and backward pass on a $n$ layer neural network.
Assume that we store $k$ intermediate results in the graph and apply the same strategy recursively when doing forward and backward pass on the sub-path.
We have the following recursion formula.
\begin{equation}
  g(n) = k + g\left(n / (k+1)\right)
\end{equation}
Solving this recursion formula gives us
\begin{equation}
  g(n) = k \log_{k+1} (n)
\end{equation}
As a special case, if we set $k = 1$, we get $g(n) = \log_2 n$. This is interesting conclusion as all the existing implementations takes $O(n)$ memory in feature map to train a $n$ layer neural network.
This will require $O(\log_2 n)$ cost forward pass cost, so may not be used commonly.
But it demonstrates how we can trade memory even further by using recursion.

\subsection{Guideline for Deep Learning Frameworks}
In this section, we have shown that it is possible to trade computation for memory  and
combine it with the system optimizations proposed in Sec~\ref{sec:comp-graph}. It is helpful for deep learning frameworks to
\begin{itemize}
\item Enable option to drop result of low cost operations.
\item Provide planning algorithms to give efficient memory plan.
\item Enable user to set the mirror attribute in the computation graph for memory optimization.
\end{itemize}
While the last option is not strictly necessary, providing such interface enables user to hack their own memory optimizers
and encourages future researches on the related directions.
Under this spirit, we support the customization of graph mirror plan and will make the source code publicly available.

\presec
\section{Experiments}
\postsec

\subsection{Experiment Setup}

We evaluate the memory cost of storing intermediate feature maps using the methods described in this paper.
We our method on top of MXNet~\cite{MXNet-whitepaper}, which statically allocate all the intermediate feature
maps before computation. This enables us to report the \emph{exact memory cost} spend on feature maps.
Note that the memory cost of parameters and temporal memory~(e.g. required by convolution) are not part of the memory cost report.
We also record the runtime total memory cost  by running training steps on a Titan X GPU.
Note that all the memory optimizations proposed in this paper gives equivalent weight gradient for training and can always be safely applied.
We compare the following memory allocation algorithms

\begin{figure}[t!]
 \centering
 \subfigure[Feature map memory cost estimation] {
   \includegraphics[width=.47\textwidth]{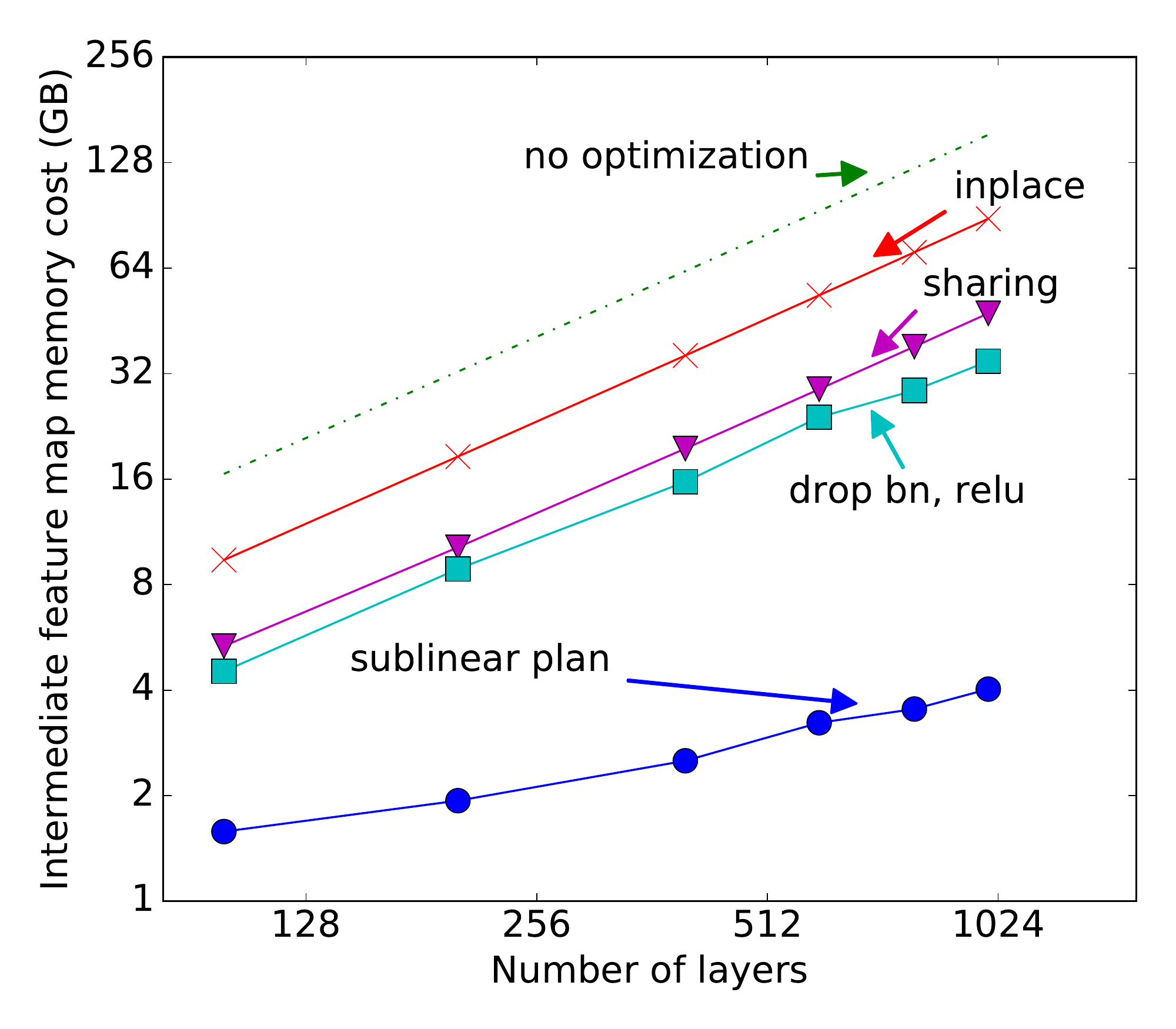}
 }
 \subfigure[Runtime total memory cost] {
   \includegraphics[width=.47\textwidth]{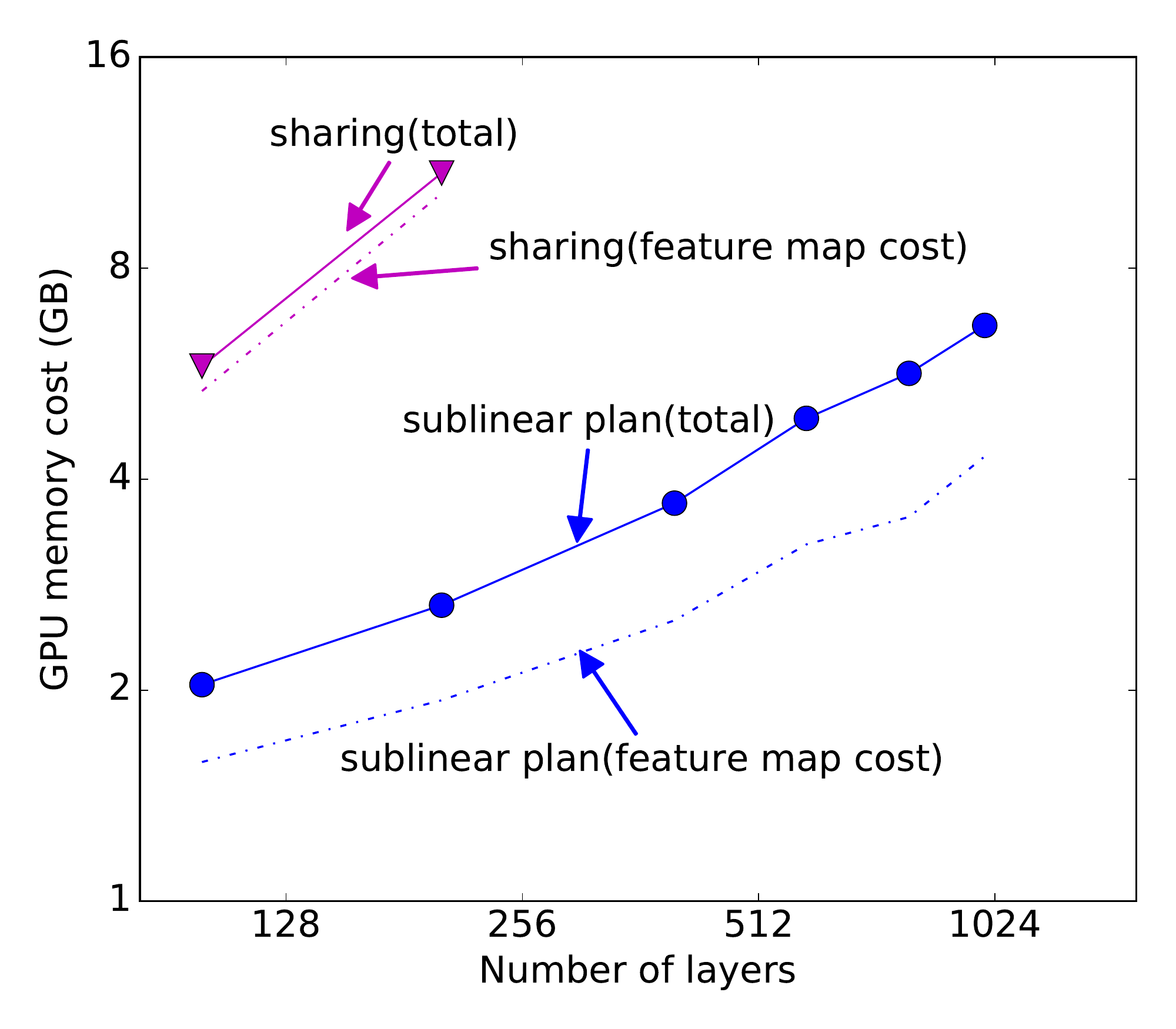}
 }
 \caption{
   The memory cost of different allocation strategies on deep residual net configurations.
   The feature map memory cost is generated from static memory allocation plan.
   We also use nvidia-smi to measure the total memory cost during runtime~(the missing points are due to out of memory).
   The figures are in log-scale, so $y = \alpha x^\beta$ will translate to $\log(y) = \beta \log(x) + \log \alpha$.
   We can find that the graph based allocation strategy indeed help to reduce the memory cost by a factor of two to three.
   More importantly, the sub-linear planning algorithm indeed gives sub-linear memory trend with respect to the workload.
   The real runtime result also confirms that we can use our method to greatly reduce memory cost deep net training.
 } \label{fig:resnet-memory}
\end{figure}

\begin{itemize}
 \item \emph{no optimization}, directly allocate memory to each node in the graph without any optimization.
 \item \emph{inplace}, enable inplace optimization when possible.
 \item \emph{sharing}, enable inplace optimization as well as sharing.
   This represents all the system optimizations presented at Sec.~\ref{sec:comp-graph}.
 \item \emph{drop bn-relu}, apply all system optimizations, drop result of batch norm and relu, this is only shown in convolutional net benchmark.
 \item \emph{sublinear plan}, apply all system optimizations, use plan search with Alg~\ref{alg:pack} to trade computation with memory.
\end{itemize}

\subsection{Deep Convolutional Network}

We first evaluate the proposed method on convolutional neural network for image classification.
We use deep residual network architecture~\cite{He2016}~(ResNet), which gives the state of art result on this task.
Specifically, we use 32 batch size and set input image shape as $(3, 224, 224)$.
We generate different depth configuration of ResNet~\footnote{We count a conv-bn-relu as one layer}
by increasing the depth of each residual stage.

We show the results in Fig.~\ref{fig:resnet-memory}.
We can find that the system optimizations introduced in Sec.~\ref{sec:comp-graph} can help to reduce
the memory cost by factor of two to three.
However, the memory cost after optimization still exhibits a linear trend with respect to number of layers.
Even with all the system optimizations, it is only possible to train a 200 layer ResNet with the best GPU we can get.
On the other hand, the proposed algorithm gives a sub-linear trend in terms of number of layers.
By trade computation with memory, we can train a 1000 layer ResNet using less than 7GB of GPU memory.

\subsection{LSTM for Long Sequences}
\begin{figure}[t!]
 \centering
 \subfigure[Feature map memory cost estimation] {
   \includegraphics[width=.47\textwidth]{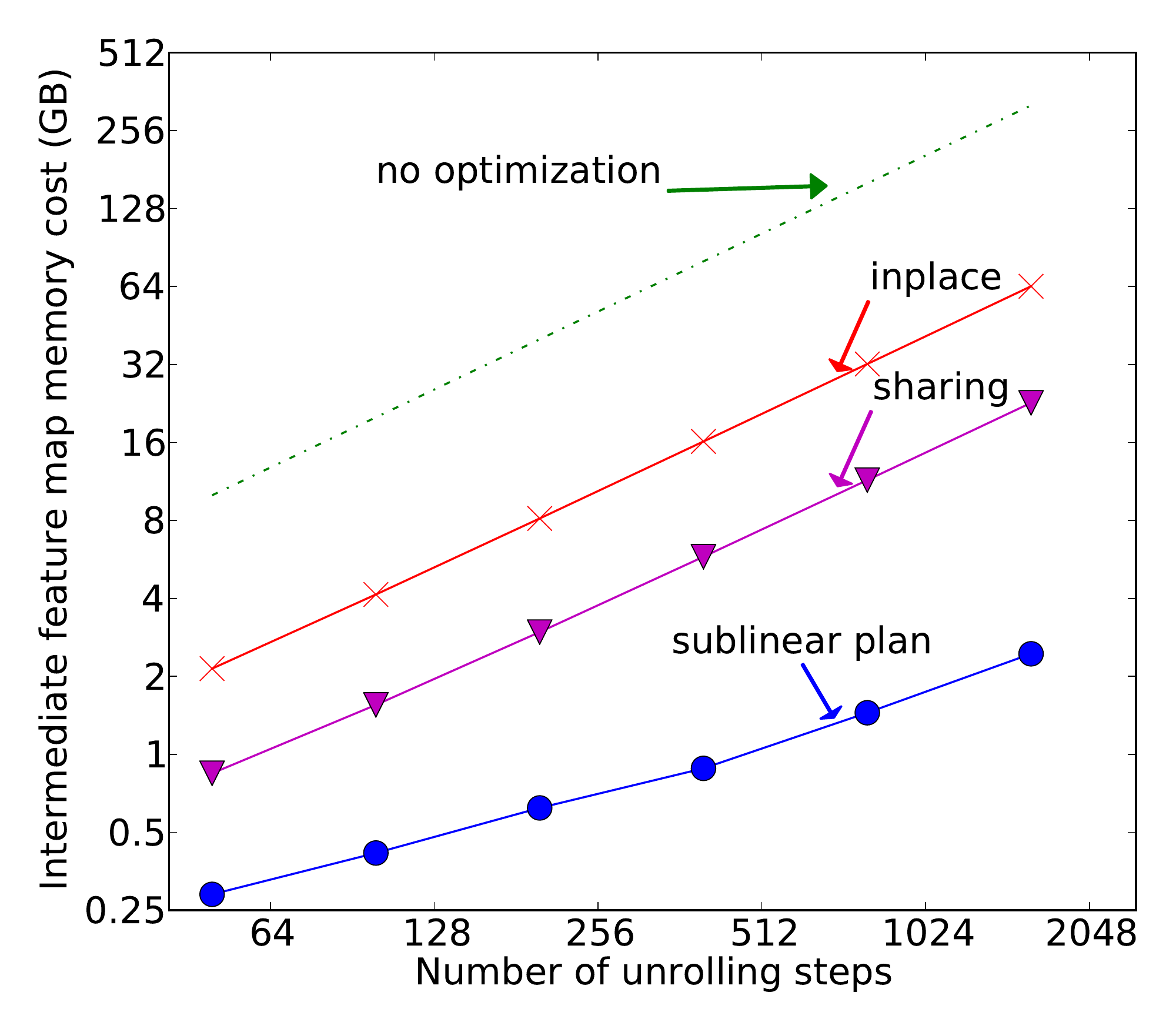}
 }
 \subfigure[Runtime total memory cost] {
   \includegraphics[width=.47\textwidth]{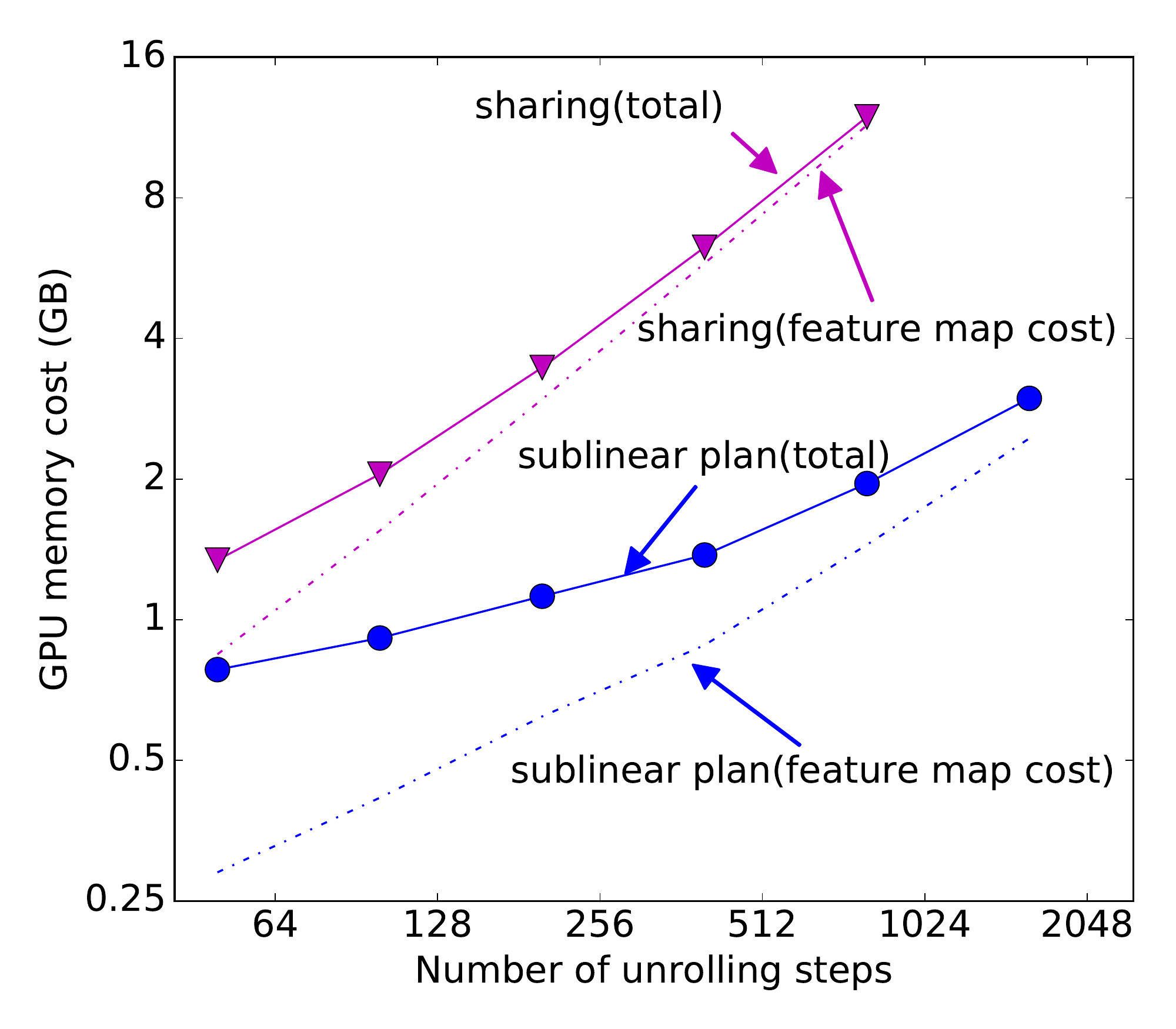}
 }
 \caption{
   The memory cost of different memory allocation strategies on LSTM configurations.
   System optimization gives a lot of memory saving on the LSTM graph, which contains
  a lot of fine grained operations. The sub-linear plan can give more than 4x reduction
   over the optimized plan that do not trade computation with memory.
 } \label{fig:lstm-memory}
\end{figure}
\begin{figure}[!t]
 \centering
 \subfigure[ResNet] {
   \includegraphics[width=.47\textwidth]{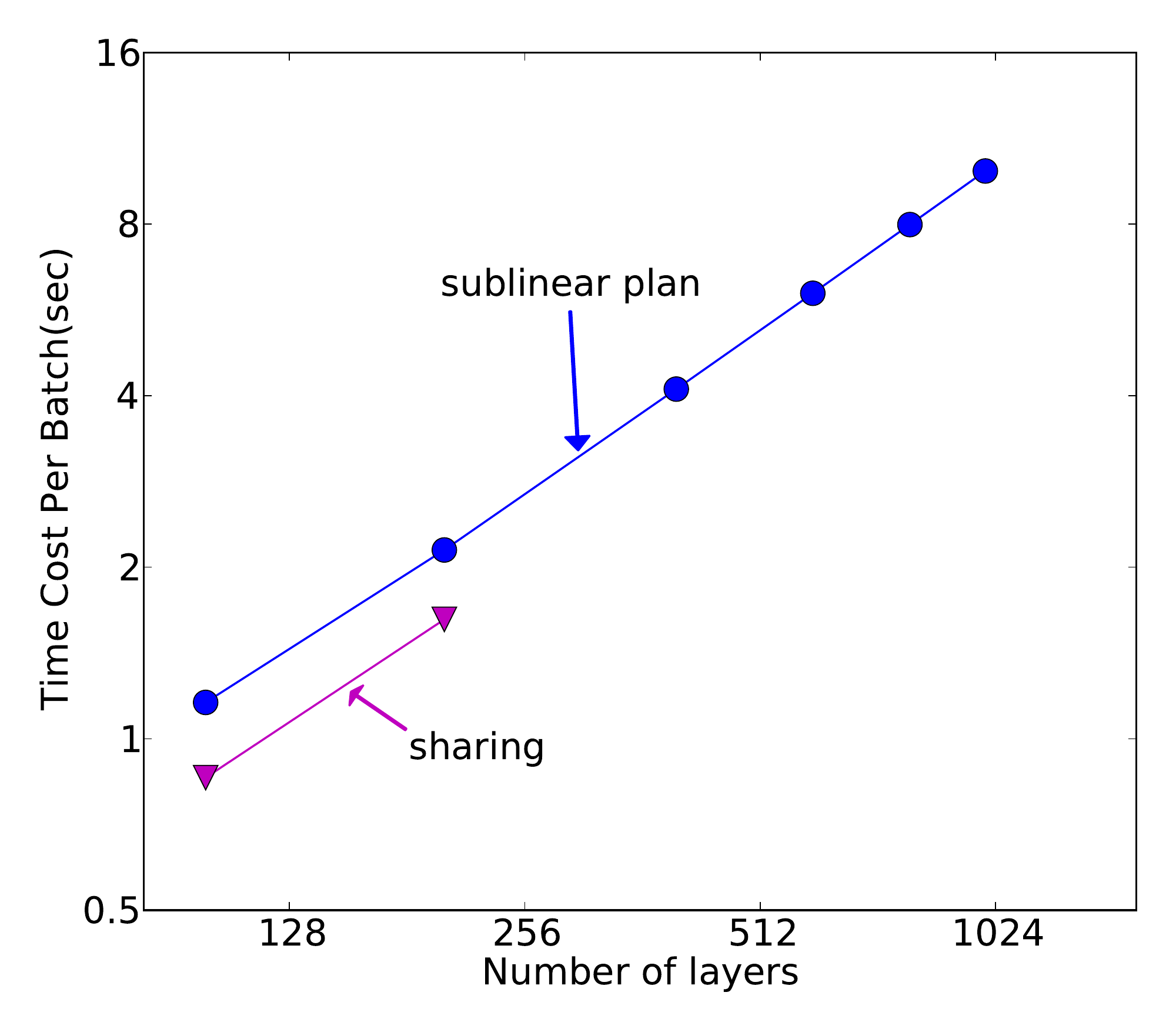}
 }
 \subfigure[LSTM] {
   \includegraphics[width=.47\textwidth]{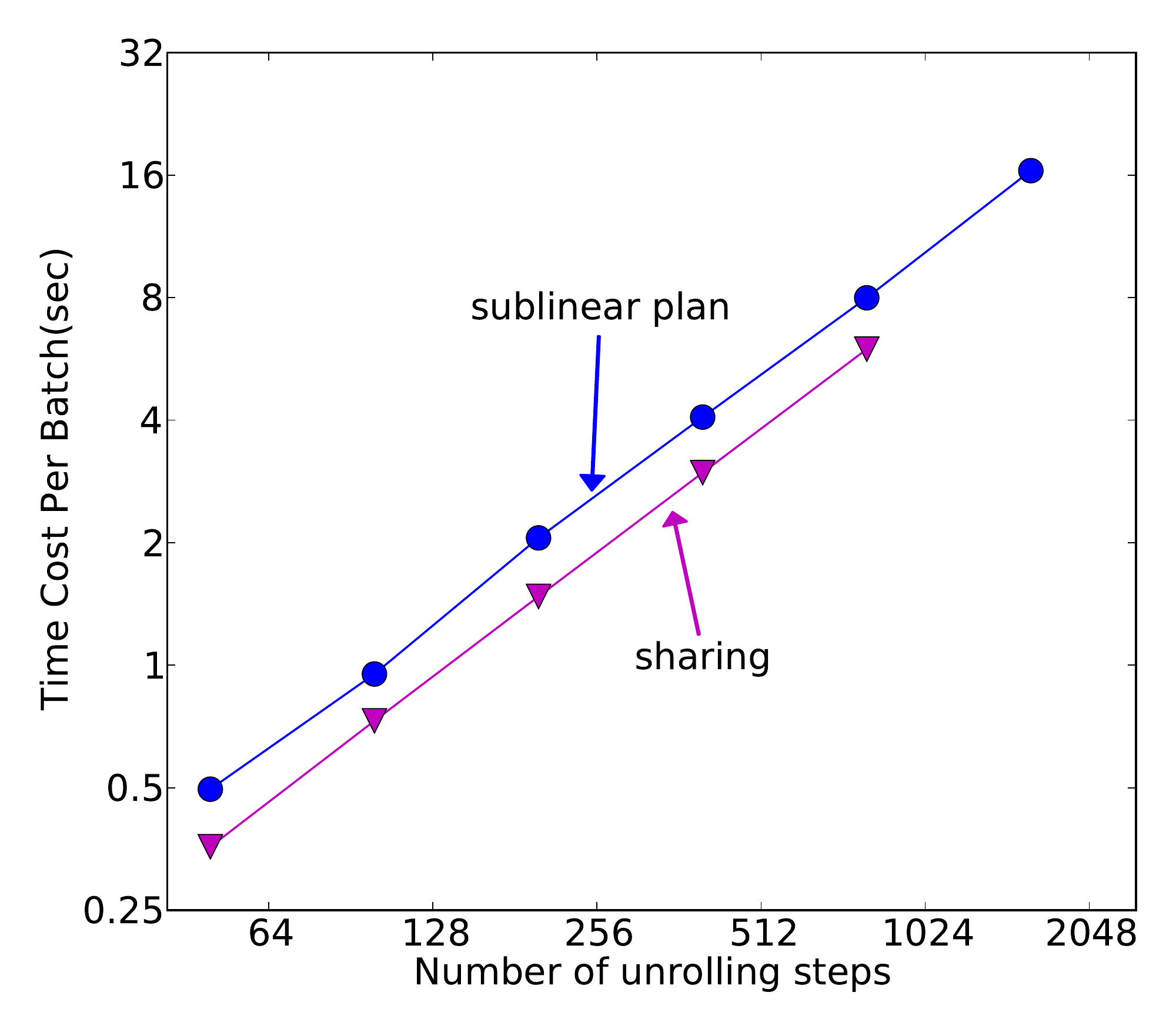}
 }
 \caption{
   The runtime speed of different allocation strategy on the two settings.
   The speed is measured by a running 20 batches on a Titan X GPU.
   We can see that using sub-linear memory plan incurs roughly $30$\% of
   additional runtime cost compared to linear memory allocation.
   The general trend of speed vs workload remains linear for both strategies.
 } \label{fig:speed}
\end{figure}
We also evaluate the algorithms on a LSTM under a long sequence unrolling setting.
We unrolled a four layer LSTM with 1024 hidden states equals 64 over time.
The batch size is set to 64.
The input of each timestamp is a continuous 50 dimension vector and the output is softmax over 5000 class.
This is a typical setting for speech recognition\cite{DBLP:conf/interspeech/SakSB14}, but our result
can also be generalized to other recurrent networks.
Using a long unrolling step can potentially help recurrent model to learn long term dependencies over time.
We show the results in Fig.~\ref{fig:lstm-memory}.
We can find that inplace helps a lot here. This is because inplace optimization in our experiment enables
direct addition of weight gradient to a single memory cell, preventing allocate space for gradient at each timestamp.
The sub-linear plan gives more than 4x reduction over the optimized memory plan.

\subsection{Impact on Training Speed}

We also measure the runtime cost of each strategy. The speed is benchmarked on a single Titan X GPU.
The results are shown in Fig.~\ref{fig:speed}.
Because of the double forward cost in gradient calculation, the sublinear allocation strategy costs $30$\% additional runtime compared to the normal strategy.
By paying the small price, we are now able to train a much wider range of deep learning models.

\presec
\section{Conclusion}
\postsec
In this paper, we proposed a systematic approach to reduce the memory consumption of the
intermediate feature maps when training deep neural networks.
Computation graph liveness analysis is used to enable memory sharing between feature maps. We also showed that we can trade the computation
with the memory. By combining the techniques, we can train a $n$ layer deep neural network with
only $O(\sqrt{n})$ memory cost, by paying nothing more than one extra forward computation per
mini-batch.

\section*{Acknowledgement}
\begin{small}
We thank the helpful feedbacks from the MXNet community and developers.
We thank Ian Goodfellow and Yu Zhang on helpful discussions on computation memory tradeoffs.
We would like to thank David Warde-Farley for pointing out the relation to gradient checkpointing.
We would like to thank Nvidia for the hardware support.
This work was supported in part by ONR (PECASE) N000141010672, NSF IIS 1258741 and the TerraSwarm Research
Center sponsored by MARCO and DARPA.
Chiyuan Zhang acknowledges the support of a Nuance Foundation Grant.
\end{small}

\bibliography{DeepMemory}
\bibliographystyle{plain}

\appendix
\section{Search over Budget B}

Alg.~\ref{alg:pack} allows us to generate an optimized memory plan given
a single parameter $B$. This algorithm relies on approximate memory estimation for faster speed.
After we get the plan, we can use the static allocation algorithm to
calculate the exact memory cost. We can then do a grid search over $B$ to find a good memory plan.

To get the setting of the grid, we first run the allocation algorithm with $B=0$, then
run the allocation algorithm again with $B=\sqrt{x y}$. Here $x$ and $y$ are the outputs from Alg.~\ref{alg:pack} in the first run.
Here $x$ is the approximate cost to store inter-stage feature maps and $y$ is the approximate cost to run each stage. $B=\sqrt{x y}$ an estimation of each stage's memory cost.
This can already give a good memory plan. We then set grid around $B = \sqrt{x y}$ to further refine the solution.

In practice, we find that using a size $6$ grid on $[B/\sqrt{2}, \sqrt{2}B]$ can already give good memory plans in the experiments.
We implemented the allocation algorithm in python without any attempt to optimize for speed. Our code costs a few seconds
to get the plans needed in the experiments.

\end{document}